\def\year{2020}\relax
\documentclass[letterpaper]{article} 
\usepackage{aaai20}  
\usepackage{times}  
\usepackage{helvet} 
\usepackage{courier}  
\usepackage[hyphens]{url}  
\usepackage{graphicx} 
\def\UrlFont{\rm}  
\usepackage{graphicx}  
\usepackage{amsmath}
\usepackage{multirow}

\newcommand{\ie}{\textit{i}.\textit{e}.}
\newcommand{\eg}{\textit{e}.\textit{g}.}

\newcommand*{\authorimg}[1]{%
  \raisebox{-0.2\baselineskip}{%
    \includegraphics[
      height=2.2\baselineskip,
      width=2.2\baselineskip,
      keepaspectratio,
    ]{#1}%
  }%
}

\title{MULE: Multimodal Universal Language Embedding}
\author{Donghyun Kim, Kuniaki Saito, Kate Saenko, Stan Sclaroff, Bryan A. Plummer \\
Boston University\\
\{donhk, keisaito, saenko, sclaroff, bplum\}@bu.edu
}
 
\begin{document}

\maketitle

\begin{abstract}
Existing vision-language methods typically support two languages at a time at most.  In this paper, we present a modular approach which can easily be incorporated into existing vision-language methods in order to support many languages. We accomplish this by learning a single shared \textit{Multimodal Universal Language Embedding} (MULE) which has been visually-semantically aligned across all languages.  Then we learn to relate MULE to visual data as if it were a single language. Our method is not architecture specific, unlike prior work which typically learned separate branches for each language, enabling our approach to easily be adapted to many vision-language methods and tasks.  Since MULE learns a single language branch in the multimodal model, we can also scale to support many languages, and \textit{languages with fewer annotations} can take advantage of the good representation learned from other (more abundant) language data. We demonstrate the effectiveness of MULE on the bidirectional image-sentence retrieval task, supporting up to four languages in a single model. In addition, we show that Machine Translation can be used for data augmentation in multilingual learning, which, combined with MULE, improves mean recall by up to 21.9\% on a single-language compared to prior work, with the most significant gains seen on languages with relatively few annotations. Our code is publicly available\footnote{\url{http://cs-people.bu.edu/donhk/research/MULE.html}}.

\end{abstract}

\section{Introduction}

Vision-language understanding has been an active area of research addressing many tasks such as image captioning~\cite{fang2015captions,gu2018unpaired}, visual question answering~\cite{AntolICCV2015,balanced_vqa_v2}, image-sentence retrieval~\cite{wang2019learning,nam2017dual}, and phrase grounding~\cite{plummer2015flickr30k,hu2015natural}. Recently there has been some attention paid to expanding beyond developing monolingual (typically English-only) methods by also supporting a second language in the same model (\eg,~\cite{gellaEMNLP2017,hitschler-etal-2016-multimodal,rajendran2015bridge,calixto2017multilingual,li2019coco,Lan:2017:FCI:3123266.3123366}).  However, these methods often learn completely separate language representations to relate to visual data, resulting in many language-specific model parameters that grow linearly with the number of supported languages.


\begin{figure}[t]
\begin{center}
   \includegraphics[width=\linewidth]{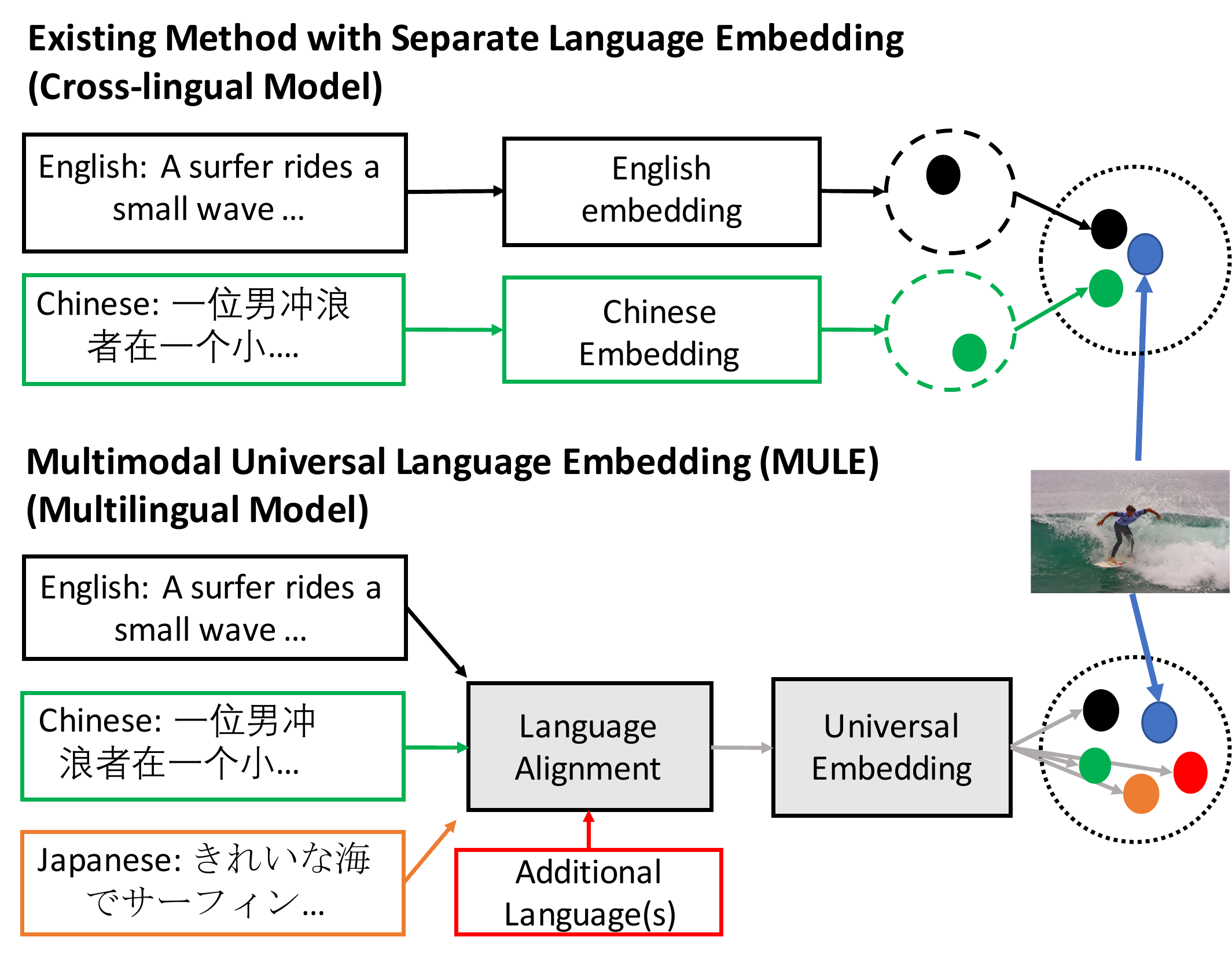}
\end{center}
   \caption{Most prior work on vision-language tasks supports up to two languages where each language is projected into a shared space with the visual features using its own language-specific model parameters (top). Instead, we propose MULE, a language embedding that is visually-semantically aligned across multiple languages (bottom).  This enables us to share a single multimodal model, significantly decreasing the number of model parameters, while also performing better than prior work using separate language branches or multilingual embeddings which were aligned using only language data.}
\label{fig:task_comparison}
\end{figure}

\begin{figure*}[t!]
\begin{center}
   \includegraphics[width=0.8\linewidth]{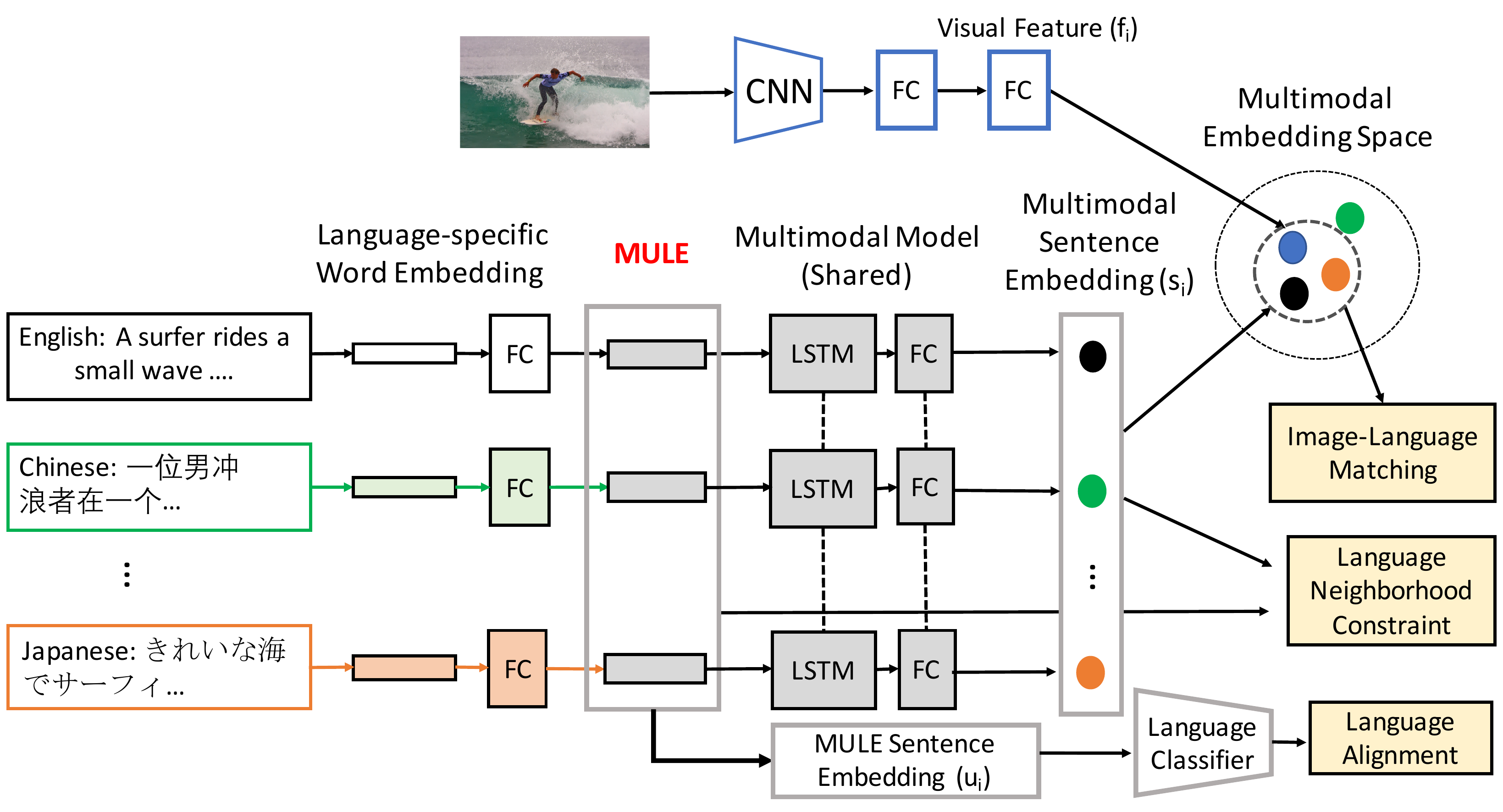}
\end{center}
   \caption{An overview of the architecture used to train our multimodal universal language embedding (MULE).  Training MULE consists of three components: neighborhood constraints which semantically aligns sentences across languages, an adversarial language classifier which encourages features from different languages to have similar distributions, and a multimodal model which helps MULE learn the visual-semantic meaning of words across languages by performing image-sentence matching.}
\label{fig:pipeline}
\end{figure*}

In this paper, we propose a Multimodal Universal Language Embedding (MULE), an embedding that has been visually-semantically aligned across many languages.  Since each language is embedded into to a shared space, we can use a single task-specific multimodal model, enabling our approach to scale to support many languages. 
 Most prior works use a vision-language model that supports at most two languages with separate language branches (\eg~\cite{gellaEMNLP2017}), significantly increasing the number of parameters compared to our work (see Fig.~\ref{fig:task_comparison} for a visualization).  
 A significant challenge of multilingual embedding learning is the considerable disparity in the availability of annotations between different languages. For English, there are many large-scale vision-language datasets to train a model such as MSCOCO~\cite{lin2014microsoft} and Flickr30K~\cite{young2014image}, but there are few datasets available in other languages, and some contain limited annotations (see Table~\ref{tab:stat} for a comparison of the multilingual datasets used to train MULE). One could simply use Neural Machine Translation (\eg~\cite{bahdanau2014neural,sutskever2014sequence}) to convert the sentence from the original language to a language with a trained model, but this has two significant limitations.  First, machine translations are not perfect and introduce some noise, making vision-language reasoning more difficult. 
Second, even with a perfect translation, some information is lost going between languages.  For example,\authorimg{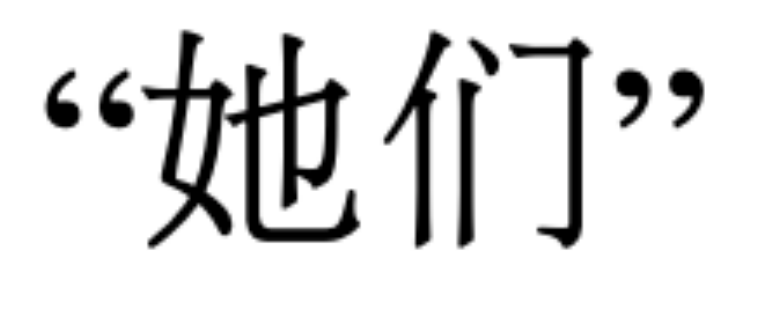} is used to refer to a group of women in Chinese.  However, it is translated to ``they'' in English, losing all gender information that could be helpful in a downstream task.  Instead of fully relying on translations, we introduce a scalable approach that supports queries from many languages in a single model.

\begin{table}[t]
\centering
\begin{tabular}{|c|c|c|c|}
\hline
Dataset   & Language & \# images & \# descriptions \\ \hline
\multirow{4}{*}{Multi30K} & English  & 29K   & 145K   \\ 
 & German   & 29K   & 145K     \\ 
 & Czech    & 29K   & 29K      \\ 
 & French   & 29K   & 29K      \\ \hline
\multirow{3}{*}{MSCOCO}   & English  & 121K  & 606K    \\ 
   & Japanese & 24K   & 122K     \\ 
   & Chinese  & 18K   & 20K     \\ \hline
\end{tabular}
\caption{Available data for each language during training.}
\label{tab:stat}
\end{table}

An overview of the architecture we use to train MULE is provided in Fig.~\ref{fig:pipeline}.  For each language we use a single fully-connected layer on top of each word embedding to project it into an embedding space shared between all languages, \ie, our MULE features.  Training our embedding consists of three components.  First, we use an adversarial language classifier in order to align feature distributions between languages.  Second, motivated by the sentence-level supervision used to train language embeddings~\cite{bert,kiela-etal-2018-learning,luVisBERT2019},
we incorporate visual-semantic information by learning how to match image-sentence pairs using a multimodal network similar to~\cite{wang2019learning}.  Third, we ensure semantically similar sentences are embedded close to each other (referred to as neighborhood constraints in Fig.~\ref{fig:pipeline}).  Since MULE does not require changes to the architecture of the multimodal model like prior work (\eg,~\cite{gellaEMNLP2017}), our approach can easily be incorporated to other multimodal models. 

Despite being trained to align languages using additional large text corpora across each supported language, our experiments will show recent multilingual embeddings like MUSE~\cite{conneau2017word} perform significantly worse on tasks like multilingual image-sentence matching than our approach. 
In addition, sharing all the parameters of the multimodal component of our network enables languages with fewer annotations to take advantage of the stronger representation learned using more data.  Thus, as our experiments will show, MULE obtains its largest performance gains on languages with less training data.  This gain is boosted further by using Neural Machine Translation as a data augmentation technique to increase the available vision-language training data. 

We summarize our contributions as follows:
\begin{itemize}
\item We propose MULE, a multilingual text representation for vision-language tasks that can transfer and learn textual representations for low-resourced languages from label-rich languages, such as English. 
\item We demonstrate MULE's effectiveness on a multilingual image-sentence retrieval task, where we outperform extensions of prior work by up to 21.9\% on the low-resourced language while also using fewer model parameters.
\item We show that using Machine Translation is a beneficial data augmentation technique for training multilingual embeddings for vision-language tasks.
\end{itemize}

\section{Related Work}


\noindent\textbf{Language Representation Learning.}
Word embeddings, such as Word2Vec~\cite{mikolov2013linguistic} and FastText~\cite{bojanowski2017enriching}, play an important role in vision-language tasks. These word embeddings provide a mapping function from a word to an n-dimensional vector where semantically similar words are embedded close to each other and are typically trained using language-only data. 
 However, recent work has demonstrated a disconnect between how these embeddings are evaluated and the needs of vision-language tasks~\cite{burnsLanguage2019}.  Thus, several recent methods have obtained significant performance gains across many tasks over language-only trained counterparts by learning the visual-semantic meaning of words specifically for use in vision-language problems~\cite{Kottur_2016_CVPR,kiela-etal-2018-learning,burnsLanguage2019,luVisBERT2019,guptaEmbeddings2019,Nguyen_2019_CVPR,tanTransformersEMNLP2019}. All these methods have addressed embedding learning only in the monolingual (English-only) setting, however, and none of the methods that align representations across many languages were designed specifically for vision-language tasks (\eg~\cite{conneau2017word,rajendran2015bridge,calixto2017multilingual}).  Thus, just as in the monolingual setting, and verified in our experiments, these multilingual, language-only trained embeddings do not generalize as well to vision-language tasks as the visually-semantically aligned multilingual embeddings in our approach.
\smallskip

\noindent\textbf{Image-Sentence Retrieval.} The goal of this task is to retrieve relevant images given a sentence query and vice versa. Although there has been considerable attention given to this task, nearly all have focused on supporting queries in a single language, which is nearly always English (\eg~\cite{nam2017dual,wang2019learning}). These models tend to either learn an embedding between image and text features (\eg~\cite{plummer2015flickr30k,wang2019learning,lee2018stacked,huangArxiv2017}) or sometimes directly learn a similarity function (\eg~\cite{wang2019learning}).  Most relevant to our work is \cite{gellaEMNLP2017} who propose a cross-lingual model, which uses an image as a pivot and enforce the sentence representations from English and German to be similar to the pivot image representation, similar to the structure-preserving constraints of \cite{wang2019learning}. However, in \cite{gellaEMNLP2017}~each language is modeled with a completely separate language model.  While this may be acceptable for modeling one or two languages, it would not scale well for representing many languages as the number of parameters would grow too large. \cite{Wehrmann_2019_ICCV} proposes a character-level encoding for a cross-lingual model, which effectively reduces the size of the word embedding for languages. However, this approach shows a significant drop in performance when training for just two languages. 

In this work we explore multiple languages with \textit{underrepresented and low-resourced languages} (up to 4 languages). We learn a shared representation between all languages, enabling us to scale to many languages with few additional parameters.  This enables feature sharing with low-resourced languages, resulting in significantly improved performance, even when learning many languages. 
\smallskip

\noindent\textbf{Neural Machine Translation.}  In Neural Machine Translation (NMT) the goal is to translate text from one language to another language with parallel text corpora~\cite{bahdanau2014neural,sutskever2014sequence,johnson2017google}. \cite{johnson2017google} proposed a multilingual NMT model, which uses a single model with an encoder-decoder architecture. They observed that translation quality on low-resourced languages can be improved when trained with label-rich languages. As discussed in the Introduction, and verified in our experiments, directly using NMT for vision-language tasks has some limitations in its usefulness for vision-language tasks, but it can provide additional benefits combined with our method.

\section{Visual-Semantic Multilingual Alignment}

In this section we describe how we train MULE, a lightweight multilingual embedding which is visually-semantically aligned across many languages and can easily be incorporated into many vision-language tasks and models. Each word in some language input is encoded using a continuous vector representation, which is then projected to the shared language embedding (MULE) using a language-specific fully connected layer.  In our experiments, we initialize our word embeddings from 300-dimensional monolingual FastText embeddings~\cite{bojanowski2017enriching}. The word embeddings and these fully connected layers are the only language-specific parameters in our network.  Due to their compact size, they can easily scale to a large vocabulary encompassing many languages.

To train MULE, we use paired and unpaired sentences between the languages from annotated vision-language datasets.  We find that we get the best performance by first pretraining MULE with paired sentences before fine-tuning using the multimodal layers with the multi-layer neighboring constraints described in (Eq.~\ref{eq:sentence_align}) and the adversarial language classifier described below. While our experiments focus solely on utilizing multimodal data, one could also try to integrate large text corpora with annotated language pairs (\eg~\cite{conneau2017word}).  However, as our experiments will show, only using generic language pairs for this alignment (\ie, not sentences related to images) results in some loss of information that is important for vision-language reasoning.  We will now discuss the three major components of our loss used to train our embedding as shown in Fig~\ref{fig:pipeline}.

\subsection{Multi-Layer Neighborhood Constraints} \label{sec:sentence_align}

During training we assume we have paired sentences obtained from the vision-language annotations, \ie, sentences that describe the same image.  These sentences are typically independently generated, so they may not refer to the same entities in the image, and when they do describe the same object they may be referenced in different ways (\eg, \emph{a black dog} vs.\ \emph{a Rottweiler}).  However, we assume they convey the same general sentiment since they describe the same image.  Thus, the multi-layer neighborhood constraints try to encourage sentences from the same image to embed near each other.  These constraints are analogous to those proposed in related work on image-sentence matching~\cite{gellaEMNLP2017,wang2019learning}, except that we apply the constraints at multiple layers of our network.  
Namely, we use the neighborhood constraints on the MULE layer as well as the multimodal embedding layer as done in prior work.  

To obtain sentence representations in the MULE space, we simply average the features of each word, which we found to perform better than using an LSTM while increasing model efficiency (an observation also made by \cite{burnsLanguage2019,wang2019learning}). For the multimodal embedding space, we use the same features for a multimodal sentence representation that is used to relate to the image features.  We denote the averaged representations in the MULE space (\ie~MULE sentence embeddings) as $u_i$  and multimodal sentence embeddings as $s_i$ as shown in Fig.~\ref{fig:pipeline}.




The neighborhood constraints are enforced using a triplet loss function. For some specific sentence embedding $s_i$, where $s_{i^{+}}$ and $s_{i^{-}}$ denote a positive and negative pair for $s_i$, respectively.  We use the same notation for positive and negative pairs $u_{i^{+}}$ and $u_{i^{-}}$. Positive and negative pairs may be from any language. So, for example, German and Czech sentences describing the same image are all positive pairs, while any pair of sentences from different images we assume are negatives (analogous assumptions were made in~\cite{gellaEMNLP2017,wang2019learning}). Given a cosine distance function $d$, the margin-based triplet loss is to minimize with a margin $m$:
\begin{equation}
\begin{aligned}
     \mathcal{L}_{LM} = \max&(0, d(s_i, s_{i^{+}}) - d(s_i, s_{i^{-}}) + m) \\
     +\max&(0, d(u_i, u_{i^{+}}) - d(u_i, u_{i^{-}}) + m). \\
\end{aligned}
\label{eq:sentence_align}
\end{equation}
\noindent Following \cite{wang2019learning}, we enumerate all positive and negative pairs in a minibatch and use the top $K$ most violated constraints, where $K=10$ in our experiments.


\subsection{Language Domain Alignment} 
\label{sec:language_classifier}

Inspired by the domain adaptation approach of \cite{ganin2014unsupervised,tzeng2014deep}, we use an adversarial language classifier (LC) to align the feature distributions of the different languages supported by our model. The goal is to project each language domain into a single shared domain, so that the model transfers knowledge between languages. This classifier does not require paired language data. We use a single fully connected layer for the LC denoted by $W_{lc}$. Given a MULE sentence representation $u_i$ presented in $l$-th language, we first minimize the objective function w.r.t the language classifier $W_{lc}$:
\begin{equation} 
\begin{aligned}
    \mathcal{L}_{LC}(W_{lc}, u_i, l) &= CrossEntropy(W_{lc} u_i, l) \\
\end{aligned}
\end{equation}
Then, in order to align the language domain, we learn language-specific parameters to maximize the loss function.



\subsection{Image-Language Matching}
\label{sec:multimodal}

To directly learn the visual meaning of words we also use a multimodal model to relate sentences to images which is trained along with our MULE embedding.  To accomplish this, we use a two-branch network similar to that of \cite{wang2019learning}, except we use the last hidden state of an LSTM to obtain a final multimodal sentence representation ($s_i$ in Fig.~\ref{fig:pipeline}).  Although \cite{burnsLanguage2019,wang2019learning} found mean-pooled features followed by a pair of fully connected layers often perform better, we found using an LSTM to be more stable in our experiments. We also kept image representation fixed, and only the two fully connected layers after the CNN in Fig.~\ref{fig:pipeline} were trained.

 Let $f_{i}$ denote the image representation and $s_i$ denote the sentence representation in the multimodal embedding space for the i-th image $x_i$. We construct a minibatch that contains positive image-sentence pairs from different images. In the batch, we get $(f_{i}, s_{i})$ from the image-sentence pair $(x_i, y_i)$. It should be noted that sentences can be presented in multiple languages. We sample triplets to have negative pairs and positive pairs for image representations and sentence representations. To be specific, given $f_{i}$, we sample corresponding positive sentence representation $s_{i^{+}}$ and a negative sentence representation $s_{i^{-}}$ represented in the same language. Equivalently, given a $y_{i}$, we sample the positive image representation $f_{i^{+}}$ and a negative image representation $f_{i^{-}}$. Then, our margin-based objective function for matching is to minimize with a margin $m$ and a cosine distance function $d$:
\begin{equation}
\begin{aligned}
     \mathcal{L}_{triplet} = & \max(0, d( f_{i^{+}}, s_{i^{+}}) - d( f_{i^{+}}, s_{i^{-}}) + m) \\
     & + \max(0, d( s_{i^{+}}, f_{i^{+}}) - d( s_{i^{+}}, f_{i^{-}}) + m).
\end{aligned}
\label{eq:bidirectional_triplet}
\end{equation}
\noindent 
As with the neighborhood constraints, the loss is computed over the $K=10$ most violated constraints.
Finally, our overall objective function is to find:
\begin{equation}
\begin{aligned}
     \hat{\theta} = &\operatorname*{argmin}_{\theta}  \lambda_1 \mathcal{L}_{LM}  - \lambda_2 \mathcal{L}_{LC} + \lambda_3 \mathcal{L}_{triplet}\\
     \hat{W_{lc}} = &\operatorname*{argmin}_{W_{lc}}  \lambda_2 \mathcal{L}_{LC}
\end{aligned}
\label{eq:total_loss}
\end{equation}
\noindent where $\theta$ includes all parameters in our network except for the language classifier, $W_{lc}$ contains the parameters of the language classifier, and $\lambda$ determines weights on each loss.

\section{Experiments}

\subsection{Datasets} 

\noindent\textbf{Multi30K}~\cite{elliott2016multi30k,elliott2017findings,barrault2018findings}. The Multi30K dataset augments Flickr30K~\cite{young2014image} with image descriptions in German, French, and Czech. Flickr30K contains 31,783 images where each image is paired with five English descriptions.  There are also five sentences provided per image in German, but only one sentence per image is provided for French and Czech.  French and Czech sentences are translations of their English counterparts, but German sentences were independently generated.  We use the dataset's provided splits which uses 29K/1K/1K images for training/test/validation.
\smallskip


\noindent\textbf{MSCOCO}~\cite{lin2014microsoft}. MSCOCO is a large-scale dataset which contains 123,287 images and each image is paired with 5 English sentences. Although this accounts for a much larger English training set compared with Multi30K, but there are fewer annotated sentences in other languages. \cite{miyazaki2016cross} released the YJ Captions 26K dataset which contains about 26K images in MSCOCO where each image is paired with independent 5 Japanese descriptions. \cite{li2019coco} provides 22,218 independent Chinese image descriptions for 20,341 images in MSCOCO. There are only about 4K image descriptions which are shared across the three languages. Thus, in this dataset, an additional challenge is the need to use unpaired language data.  We randomly selected 1K images for the testing and validation sets from the images which contain descriptions across all three languages, for a total of 2K images, and used the rest for training. Since we use the different data split, it is not possible to compare directly with prior monolingual methods. We provide a fair comparison with our baseline and prior monolingual methods in the supplementary.
\smallskip

\noindent\textbf{Machine Translations.} As shown in Table~\ref{tab:stat}, there is considerable disparity in the availability of annotations for training in different languages. As a way of augmenting these datasets, we use Google's online translator to generate sentences in other languages. Since the sentences in other languages are independently generated, their translations can provide additional variation in the training data. This also enables us to evaluate the effectiveness of NMT. In addition, we use these translated sentences to benchmark the performance translating languages from an unsupported language into one of the languages for which we have a trained model (\eg~translate a sentence from Chinese into English and perform the query using an English-trained model).
\smallskip



\subsection{Image-Sentence Matching Results}

\noindent\textbf{Metrics.}  
Performance on the image-sentence matching task is typically reported as Recall$@K=[1, 5, 10]$ for both image-to-sentence and sentence-to-image (\eg~as done in~\cite{gellaEMNLP2017,nam2017dual,wang2019learning}), resulting in performance reported over six values per language. Results reporting performance over all the six values for each language can be found in the supplementary. In this paper, we average them to obtain an overall score (mR) for each compared method/language. 
\smallskip

\noindent\textbf{Model Architecture.}  We compare the following models:

\begin{itemize}
    \item \textbf{EmbN}~\cite{wang2019learning}.  As shown in \cite{burnsLanguage2019}, EmbN is the state-of-the-art image-sentence model when using image-level ResNet features and good language features.  This model is the multimodal network in Fig.~\ref{fig:pipeline}. 

    \item \textbf{PARALLEL-EmbN.} This model borrows ideas from \cite{gellaEMNLP2017} to modify EmbN.  Specifically, only a single image representation is trained, but it contains separate language branches.
\end{itemize}

\noindent\textbf{Multi30K Discussion.}
We report performance on the Multi30K dataset in Table~\ref{tab:multi30k}. 
The first line of Table~\ref{tab:multi30k}(a) reports performance when training completely separate models (\ie~no shared parameters) for each language in the dataset.  The significant discrepancy between the performance of English and German compared to Czech and French can be attributed to the differences in the number of sentences available for each language (Czech and French have 1/5th the sentences as seen in Table~\ref{tab:stat}). Performance improves across all languages using the PARALLEL model in Table~\ref{tab:multi30k}(a), demonstrating that the representation learned for the languages with more available annotations can still be leveraged to the benefit of other languages.


Table~\ref{tab:multi30k}(b) and Table~\ref{tab:multi30k}(c) show the the results of using multilingual embeddings, ML BERT~\cite{bert} and MUSE~\cite{conneau2017word} which learns a shared FastText-style embedding space for all supported languages. This enables us to compare against aligning languages using language-only data vs.\ our approach which performs a visual-semantic language alignment.  Note that a single EmbN model is trained across all languages when using MUSE rather than training separate models since the embeddings are already aligned across languages. Comparing the numbers of Table~\ref{tab:multi30k}(a) and Table~\ref{tab:multi30k}(b), we observe that ML BERT which is a state-of-the-art method in NLP performs much worse than the monolingual FastText. In addition, we see in Table~\ref{tab:multi30k}(c) that MUSE improves performance on low-resourced languages (\ie~French and Czech), but actually hurts performance on the language with more available annotations (\ie~English). These results indicate that some important visual-semantic knowledge is lost when relying solely on language-only data to align language embeddings and NLP method does not generalize well to the language-vision task.

Table~\ref{tab:multi30k}(d) compares the effect that different components of MULE has on performance.
Going from the last line of Table~\ref{tab:multi30k}(a) to the first line of Table~\ref{tab:multi30k}(d) demonstrates that using a single-shared language branch can significantly improve lower-resource language performance (\ie~French and Czech), with only a minor impact to performance on languages with more annotations.  Comparing the last line of Table~\ref{tab:multi30k}(c) which reports performance of our full model using MUSE embeddings, to the last line of Table~\ref{tab:multi30k}(d), we see that using MUSE embeddings still hurts performance, which helps verify our earlier hypothesis that some important visual-semantic information is lost when aligning languages with only language data.  This is also reminiscent of an observation in~\cite{burnsLanguage2019}, \ie, it is important to consider the visual-semantic meaning of words when learning a language embedding for vision-language tasks.

Breaking down the components of our model in the last three lines of Table~\ref{tab:multi30k}(d), we show that including the multi-layer neighborhood constraints (NC), language classifier (LC), and pretraining MULE (LP) all provide significant performance improvements (a full ablation study can be found in the supplementary).  In fact, they can make up for much of the lost performance on the high-resource languages when sharing a single language branch in the multimodal model, with German actually outperforming its separate language-branch counterpart. French and Czech perform even better, however, with a total improvement of 4.4\% and 7.1\% mean recall over our reproductions of prior work, respectively. Clearly, training multiple languages together in a single model, especially those with fewer annotations, can result in dramatic improvements to performance without having to sacrifice the performance of a single language as long as some care is taken to ensure the model learns a comparable representation between languages. Our method achieves the best performance on German, French, and Czech, while still being comparable for English.
\smallskip

\begin{table}[t]
\setlength{\tabcolsep}{2pt}
\centering
\begin{tabular}{|rl|c|c|c|c|c|}
\hline
&  & Single & \multicolumn{4}{|c|}{Mean Recall}\\
\cline{4-7}
& Model &  Model & En & De & Fr & Cs\\
\hline
\hline
{\bf (a)} & \textbf{FastText (Baseline)} & & & &  &\\
& EmbN & N & \textbf{71.1} & 57.9 & 43.4 & 33.4\\
& PARALLEL-EmbN & Y & 69.6 & \textbf{61.6} & 52.0 & 43.2\\
\hline
{\bf (b)} & \textbf{ML BERT} & & & &  &\\
& EmbN & Y & 45.5 & 37.9 & 36.4 & 19.2\\
& PARALLEL-EmbN & Y & 60.4 & 51.1 & 42.0 & 29.8\\
\hline
{\bf (c)} & \textbf{MUSE} & & & &  &\\
& EmbN & Y & 68.6 & 58.2 & 54.0 & 41.8\\
& PARALLEL-EmbN & Y & 69.5 & 59.0 & 51.6 & 40.7\\
& EmbN+NC+LC+LP& Y & 69.0 & 59.7 & 53.6 & 41.0\\
\hline
{\bf (d)} & \textbf{MULE (Ours)} & & & & &\\
& EmbN & Y & 67.5 & 59.2 & 52.5 & 43.9\\
& EmbN+NC & Y & 67.2 & 59.9 & 54.0 & 46.4\\
& EmbN+NC+LC  & Y & 67.1 & 60.9 & 55.5 & 49.5\\
& EmbN+NC+LC+LP (Full) & Y & 68.0 & 61.4 & \textbf{56.4} & \textbf{50.3}\\
\hline
\end{tabular}
\caption{Performance comparison of different language embeddings on the image-sentence retrieval task on Multi30K. MUSE and MULE are multilingual FastText embeddings w/ and w/o visual-semantic alignment, respectively. We denote NC: multi-layer neighborhood constraints, LC: language classifier, and LP: pretraining MULE.
}
\label{tab:multi30k}
\end{table}

\noindent\textbf{MSCOCO Discussion.}
Table~\ref{tab:coco} reports results on MSCOCO.  Here, the lower resource language is Chinese, while English and Japanese both have considerably more annotations (although, unlike German on Multi30K, English has considerably more annotations than Japanese on this dataset).  For the most part we see similar behavior on the MSCOCO dataset that we saw on Multi30K - the lower resource languages (Chinese) performs worse overall compared to the higher resource languages, but most of the performance gap is reduced when using our full model. Overall, our formulation obtains a 5.9\% improvement to mean recall over our baselines for Chinese, and also improves performance by 1.6\% mean recall for Japanese. However, for English, we obtain a slight decrease in performance compared with the English-only model reported on the first line in Table~\ref{tab:coco}(a).  

The drop in performance on English could be due to the significant imbalance in the training data on this dataset, where more than 3/4 of the data contains only English captions.  In our experiments we separated the data into three groups: English only, English-Japanese, and English-Chinese.  We ensured each group was equally represented in the minibatch, which means some images containing Japanese or Chinese captions were sampled far more than many of the English-only images.  This shift in the distribution of the training data may account for some of the loss of performance.  We believe more sophisticated sampling strategies may help rectify these issues and re-gain the lost performance. That said, our model has significantly fewer parameters from learning a single language branch for all languages while also outperforming the PARALLEL model from prior work which learns separate language branches.

\begin{table}[t]
\setlength{\tabcolsep}{2pt}
\centering
\begin{tabular}{|rl|c|c|c|c|}
\hline
& & Single & \multicolumn{3}{|c|}{Mean Recall}\\
\cline{4-6}
& Model &  Model & En & Cn & Ja\\
\hline
\hline
{\bf (a)} & \textbf{FastText (Baseline)} & & & &\\
& EmbN & N &  \textbf{75.6} & 55.7 & 69.4\\
& PARALLEL-EmbN & Y & 70.0 & 52.9 & 68.9\\
\hline
\textbf{(b)} & \textbf{ML BERT} & & & &\\ 
& EmbN & Y & 59.4 & 44.7 & 47.2\\
& PARALLEL-EmbN & Y & 57.6 & 57.5 & 62.1\\
\hline
{\bf (c)} & \textbf{MULE (Ours)} & & & &\\
& EmbN & Y & 69.4 & 54.2 & 69.0\\
& EmbN+NC & Y & 69.8 & 56.6 & 69.5\\
& EmbN+NC+LC  & Y & 71.3 & 57.9 & 70.3\\
& EmbN+NC+LC+LP (Full) & Y & 72.0 &  \textbf{58.8} &  \textbf{70.5}\\
\hline
\end{tabular}
\caption{Performance comparison of different language embeddings on the image-sentence retrieval task on MSCOCO. 
}
\label{tab:coco}
\end{table}

\begin{table*}[t]
\setlength{\tabcolsep}{2.4pt}
\centering
\begin{tabular}{|rl|c|c|c|c|c|c|c|c|c|}
\hline
&&&  & \multicolumn{4}{|c|}{Multi30K} & \multicolumn{3}{|c|}{MSCOCO}\\
\cline{5-11}
&& & Single & \multicolumn{7}{|c|}{Mean Recall}\\
\cline{5-11}
& Model  & Training Data Source &  Model & En & De & Fr & Cs & En & Cn & Ja\\
\hline
\hline
\textbf{(a)} & PARALLEL-EmbN  & Human Generated Only (Tables \ref{tab:multi30k}\&\ref{tab:coco}) & Y & 69.6 & 61.6 & 52.0 & 43.2 & 70.0 & 52.9 & 68.9 \\
&MULE EmbN - Full  & Human Generated Only (Tables \ref{tab:multi30k}\&\ref{tab:coco}) & Y & 68.0 & 61.4 & 56.4 & 50.3 & 72.0 & 58.8 & 70.5 \\
\hline
\hline
\textbf{(b)}& EmbN \& Machine & \multirow{2}{*}{Human Generated English Only} & \multirow{2}{*}{Y} & \multirow{2}{*}{71.1} & \multirow{2}{*}{48.5} & \multirow{2}{*}{46.7} & \multirow{2}{*}{46.9} & \multirow{2}{*}{75.6} & \multirow{2}{*}{72.2} & \multirow{2}{*}{66.1}\\
& Translated Query  & & & & & & & & &\\
& EmbN  & Human Generated + Machine Translations & N & \textbf{72.0} & 60.3 & 54.8 & 46.3 &  76.8 & 73.5 & 73.2\\
&PARALLEL-EmbN  & Human Generated + Machine Translations & Y & 69.0 & 62.6 & 60.6 & 54.1 & 78.3 & 73.5 & 76.0\\
&MULE EmbN - Full & En $\rightarrow$ Others, Machine Translations Only & Y & 69.3 & 62.1 & 61.5 & 55.5 & 77.3 & 73.3 & 75.3\\
&MULE EmbN - Full & Human Generated + Machine Translations & Y & 70.3 &  \textbf{64.1} &  \textbf{62.3} &  \textbf{57.7} & \textbf{79.5} &  \textbf{74.8} &  \textbf{76.3}\\
\hline
\end{tabular}
\caption{Image-sentence matching results with Machine Translation data. We translate sentences between English and the other languages (\eg~En $\longleftrightarrow$ Ja and En $\longleftrightarrow$ Cn for MSCOCO) and augment our training set with these translations.}
\label{tab:translated}
\end{table*}

\subsection{Leveraging Machine Translations}
\label{sec:machine_results}
As mentioned in the introduction, an alternative for training a model to support every language would be to use Neural Machine Translation to convert a query sentence from an unsupported language into a language which there is a trained model available.  We test this approach using an English-trained EmbN model whose performance is reported on the first lines of Table~\ref{tab:multi30k}(a) and Table~\ref{tab:coco}(a). For each non-English language, we use Google Translate to convert the sentence from the source language into English, then use an English EmbN model to retrieve the images in the test set.

The first row of Table~\ref{tab:translated}(b) reports the results of translating non-English queries into English and using the English-only model. On the Multi30K test set we see this performs worse on each non-English language than our MULE approach, but it does outperform some of the baselines trained on human-generated captions.  Similar behavior is seen on the MSCOCO data, with Chinese-translated sentences actually performing nearly as well as human-generated English sentences. In short, using translations performs better on low-resourced languages (French, Czech, and Chinese) than the baselines. These results suggest that these translated sentences are able to capture enough information from the original language to still provide a representation that is ``good enough'' to be useful.

Since translations provide a good representation for performing the retrieval task, they should also be useful in training a new model.  This is especially true for any sentences that were independently generated, as they might provide a novel sentence after being translated into other languages.  We report the performance of using these translated sentences to augment our training set for both datasets in Table~\ref{tab:translated}(b), where our model obtains best overall performance. We observe that the models with the augmentation (\eg~last line of Table~\ref{tab:translated}(b)) always outperform the corresponding models without the augmentation (\eg~last line of Table~\ref{tab:translated}(a)) on all languages. On the second line of Table~\ref{tab:translated}(b) we see that these translations are useful in providing more training examples even for a monolingual EmbN model. Comparing the fourth and last lines of Table~\ref{tab:translated}(b) we see the difference between training the non-English languages using translated sentences alone and training with both human-generated and translated sentences.  Even though the human-generated Chinese captions account for less than 5\% of the total Chinese training data, we still see a significant performance improvement using them, with similar results on all other languages.  This suggests that human-generated captions still provide better training data than machine translations. We also see comparing our full model to the PARALLEL-EmbN model and when using MUSE embeddings that using MULE provides performance benefits even when data is more plentiful.

\begin{figure}[t!]
\begin{center}
  \includegraphics[width=0.98\linewidth]{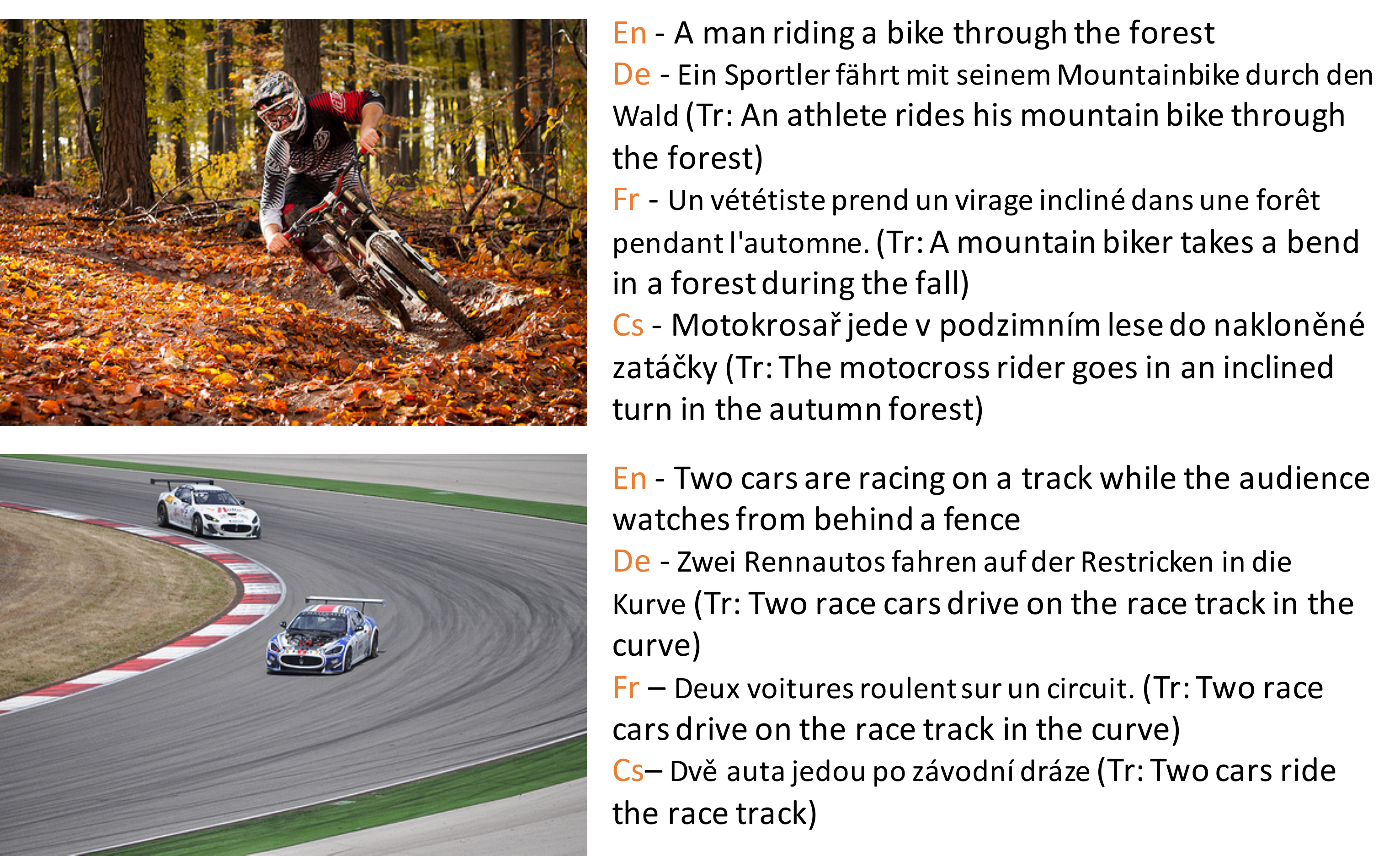}
\end{center}
  \caption{Examples of image-sentence matching results. Given an image, we pick the closest sentences on Multi30K.}
\label{fig:qaulex}
\end{figure}

\subsection{Parameter Comparison} The language branch in our experiments contained 6.8M parameters.  This results in $6.8\text{M} \times 4=27.2\text{M}$ parameters for the PARALLEL-EmbN model proposed by \cite{gellaEMNLP2017} on Multi30K (a branch for each language). MULE uses a FC layer containing $1.7\text{M}$ parameters to project word features into the universal embedding, so an EmbN model for Multi30K that uses MULE would have $6.8\text{M} + 1.7\text{M} \times 4 = 13.6\text{M}$ parameters, \textit{half the number used by \cite{gellaEMNLP2017}}. MULE also scales better with more languages than \cite{gellaEMNLP2017}. ML BERT is much larger than MULE, consisting of 12 layers with $\approx 110$M parameters. 

\subsection{Qualitative Results}
Fig.~\ref{fig:qaulex} shows the qualitative results on our full model. We pick the two samples and retrieve the closest sentences given an image for each language on Multi30K. For other languages, we provide English translations using Google Translate. The top example shows the perfect matching between the languages. The bottom image shows that the model overestimates contextual information from the image in the English sentence. It captures not only the correct event (car racing) but also wrong objects not presented in the image (audience and fence). This sentence came from similar images with minor differences in the test set. However, the minor differences in images can be important for matching between similar images. Learning how to accurately capture the details of an image may improve the performance in future work. More results can be found in supplementary.

\section{Conclusion}
We investigated bidirectional image-sentence retrieval in a multilingual setting. We proposed MULE, which can handle multiple language queries with negligible language-specific parameters unlike prior work which learned completely distinct representations for each language. In addition to being more scalable, our method enables the model to transfer knowledge between languages, resulting in especially good performance on lower-resource languages. In addition, in order to overcome limited annotations, we show that leveraging Neural Machine Translation to augment a training dataset can significantly increase performance for training both a multilingual network as well as monolingual model.  Although our work primarily focused on image-sentence retrieval, our approach is modular and can be easily incorporated into many other vision-language models and tasks. 

\section{Acknowledgements}
 This work is supported in part by Honda and by DARPA and
NSF awards IIS-1724237, CNS-1629700, CCF-1723379.
{\small
\bibliographystyle{aaai}
\bibliography{egbib}
}

\section{Implementation Details}  We train our models for 20 epochs using Adam~\cite{adam} with a learning rate of $1e^{-4}$ that we decay exponentially with a batch size of 450 images.  After obtaining the 300-dimensional FastText embeddings~\cite{bojanowski2017enriching}, they are projected into a 512-dimensional universal embedding.  Then, the universal embedding features are fed into an LSTM with 1024 units before being projected into the final 512-dimensional multimodal embedding space.  We extract our image representation using a 152-layer ResNet~\cite{He2015} that was trained on ImageNet~\cite{imagenet_cvpr09}.  An image representation was averaged over 10 crops with input image dimensions of 448x448, resulting in a 2048-dimensional image representation. After obtaining out 2048-dimensional image features, we use a pair of fully connected layers with output sizes of 2048 and 512, respectively, to project the image features into the shared multimodal embedding space.  These fully connected layers are separated with a ReLU non-linearity and use batch normalization~\cite{Ioffe:2015:BNA:3045118.3045167}.  Our language classifier is implemented as a single fully connected layer that takes mean-pooled universal embedding features as an input. We use a gradient reversal layer~\cite{ganin2014unsupervised} to implement adversarial learning on the language classifier. We set $\lambda_{1,3} = 1$ and $\lambda_{2} = 1e^{-6}$ from Eq.4.  For all distance computations, we use cosine distance.  While we keep the ResNet fixed in our experiments, we fine-tune the FastText embeddings during training.  As done in~\cite{burnsLanguage2019}, we $L2$ regularize the word embeddings to help avoid catastrophic forgetting using a regularization weight of $5e^{-7}$.
\smallskip

\section{Minibatch Construction}
In this section we explain how we set up a batch for training. We use a batch size of different 450 images for both Multi30K and MSCOCO. For Multi30K, all sentences can be paired with each language. However, there are five sentences per image for English and German, but only one sentence per image for Czech and French. Therefore, given an image, we randomly choose two sentences for English and German but one sentence for Czech and French (for a total of six sentences per image). For MSCOCO, the number of availability of different languages during training is significantly unbalanced. English has 606K sentences for 121K images, Japanese has 122K sentences for 26K images, and Chinese has 20K sentences for 18K images. This results in five sentences per image for every image containing English and Japanese, but only 1-2 sentences per image for Chinese. In addition, only about 4K images are shared (paired) across all three languages. We separate the data into three groups: English only, English-Japanese, and English-Chinese. We sample images from each group equally. Given an image, we randomly choose two sentences for English and Japanese but only select one sentence for Chinese. As a reminder, for all triplet loss functions (\ie, the multi-layer neighborhood constraints and image-sentence matching loss), we enumerate all possible triplets in a minibatch and keep at most 10 triplets with the highest loss.

\begin{table*}[h]

\centering
\setlength{\tabcolsep}{1.25pt}
\begin{tabular}{|rl|c|c|c|c|c|c|c|c|c|c|c|c|c|c|c|}
\hline
& & & \multicolumn{7}{|c|}{English} & \multicolumn{7}{|c|}{German}\\
\cline{4-17}
& & Single & \multicolumn{3}{|c|}{Image-to-Sentence} & \multicolumn{3}{|c|}{Sentence-to-Image} & & \multicolumn{3}{|c|}{Image-to-Sentence} & \multicolumn{3}{|c|}{Sentence-to-Image} &\\
\cline{4-17}
& Method & Model & R@1 & R@5 & R@10 & R@1 & R@5 & R@10 & mR & R@1 & R@5 & R@10 & R@1 & R@5 & R@10 & mR\\
\hline
\hline
{\bf (a)} & \textbf{State-of-the-art (VGG)}&  & & & & & & & & & & & & & &\\
& VSE & N & 31.6 & 60.4 & 72.7 & 23.3 & 53.6 & 65.8 & 51.2 & 29.3 & 58.1 & 71.8 & 20.3 & 47.2 & 60.1 & 47.8\\
& Order Embeddings & N & 34.8 & 63.7 & 74.8 & 25.8 & 56.5 & 67.8 & 53.9 & 26.8 & 57.5 & 70.9 & 21.0 & 48.5 & 60.4 & 47.5\\
& PARALLEL-SYM & Y  & 31.7 & 62.4 & 74.1 & 24.7 & 53.9 & 65.7 & 52.1 & 28.2 & 57.7 & 71.3 & 20.9 & 46.9 & 59.3 & 47.4\\
& PARALLEL-ASYM & Y & 31.5 & 61.4 & 74.7 & 27.1 & 56.2 & 66.9 & 53.0 & 30.2 & 60.4 & 72.8 & 21.8 & 50.5 & 62.3 & 49.7\\
& EmbN (our implementation) &  N  & 39.7 & 69.9 & 78.8 & 31.2 & 62.7 & 72.7 & 59.2 & -- & -- & -- & -- & -- & -- & --\\
\hline
{\bf (b)} & \textbf{Baselines (ResNet-152)} & & & & & & & & & & & & & & &\\
& EmbN &  N & \textbf{58.3} & \textbf{82.9} & \textbf{90.4} & \textbf{41.7} & \textbf{72.0} & \textbf{81.2} & \textbf{71.1} & 41.1 & 73.4 & 82.3 & 28.8 & 56.0 & 66.1 & 57.9\\
& PARALLEL-EmbN (2 Lang)& Y & 51.8 & 81.0 & 88.3 & 40.2 & 70.3 & 80.7 & 68.7 & \textbf{45.6} & \textbf{75.6} & \textbf{84.5} & \textbf{33.1} & \textbf{61.6} & 71.9 & \textbf{62.0}\\
& PARALLEL-EmbN & Y & 55.9 & 82.5 &  88.4 & 40.3 & 70.1 & 80.2 & 69.6 & 45.5 & 74.8 & 83.5 & 32.4 & 61.3 & \textbf{72.4} & 61.6\\
\hline
\end{tabular}
\caption{Image-sentence matching results on Multi30K. \textbf{(a)} compares models reported using VGG features that were used in prior work, \textbf{(b)} provides our adaptations of prior work as baselines using ResNet-152 features. PARALLEL-EmbN (2 Lang) represents a model trained on English and German only instead of the four languages.}
\label{tab:vgg}
\end{table*}

\section{Multiglingual BERT}
In order to compute the sentence-level representation from the multilingual BERT~\cite{bert}, we use the publicly available pretrained model and use the public API ``bert-as-service'', which takes a mean-pooling strategy for a sentence embedding. After we compute the sentence-level embedding, we cache the features and use these features to training.  Although we might be able to improve performance by fine-tuning the ML BERT model, its large size ($\approx 110$M parameters) makes it impossible to fit into GPU memory with the very large number of image-sentence pairs and additional model parameters used for training.

\begin{figure}[t]
\begin{center}
  \includegraphics[width=0.98\linewidth]{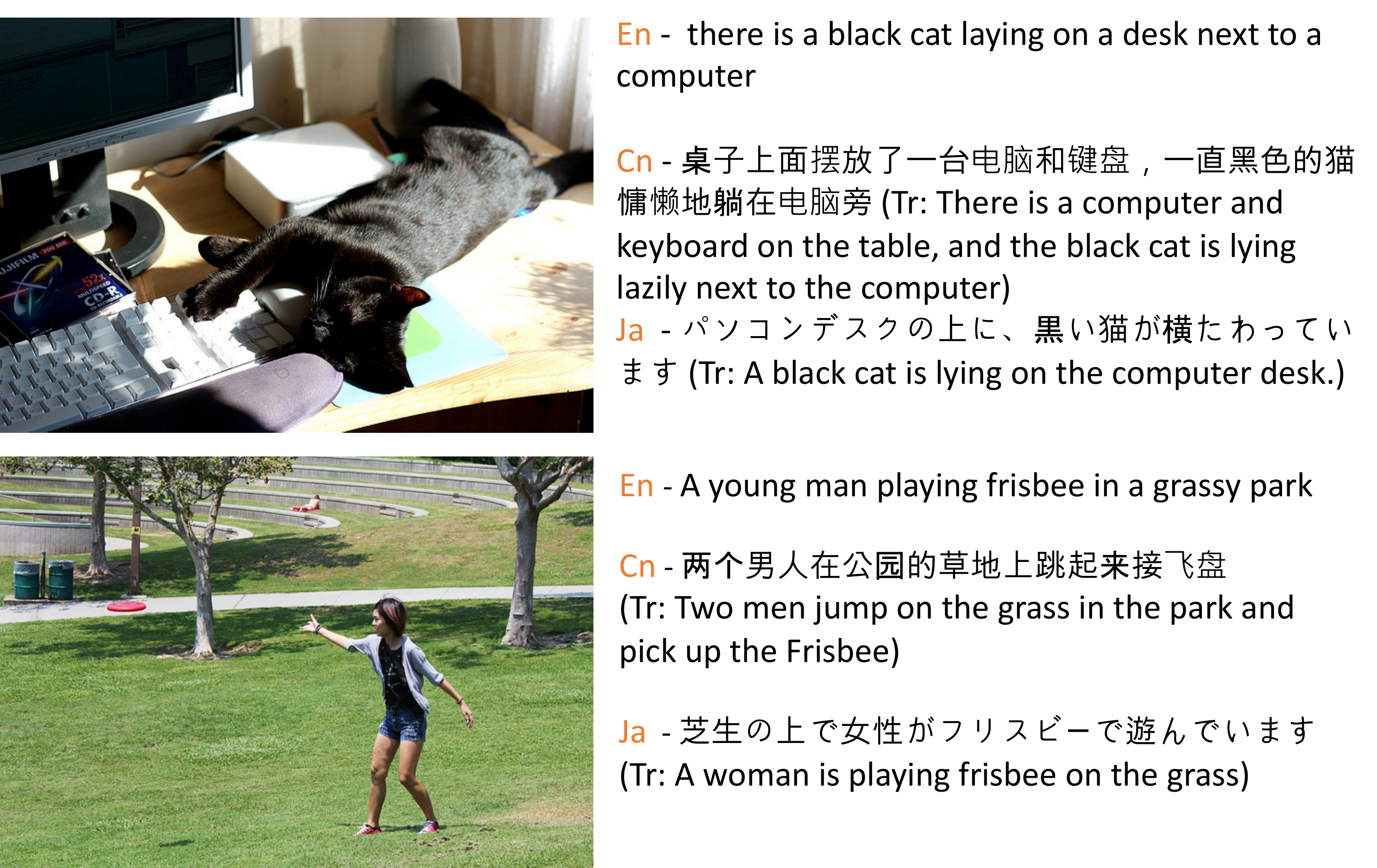}
\end{center}
  \caption{Examples of image-sentence matching results. Given an image, we pick the closest sentences on MSCOCO.}
\label{fig:qaulex_2}
\end{figure}
\section{Qualitative Results on MSCOCO}
Fig.~\ref{fig:qaulex_2} shows the qualitative results on our full model. The first example (top) shows perfect matching. In the second example (bottom), the model does not properly capture some of the the details. The subjects in the sentences are all different (\ie man vs.\ two men vs.\ a woman). These sentences all came from similar images with minor differences in the test set. However, the minor differences in images can be important for matching between similar images. Learning how to accurately capture the details of an image may improve the performance in the future work.  

\section{Detailed Analysis}
In the main paper, we report mean recall (mR) which is an average score of Recall$@1$, Recall$@5$, and Recall$@10$ on Image-Sentence retrieval. In the supplementary material, we report all the scores at each threshold on Multi30K~\cite{elliott2016multi30k,elliott2017findings,barrault2018findings} and MSCOCO~\cite{lin2014microsoft}. The scores include Image-to-Sentence and Sentence-to-Image retrieval results. Our key observations are as follows: (1) Our model allows low-resource languages to transfer knowledge from other languages while being more scalable than baselines; (2) MULE performs better than MUSE and ML-BERT in most cases; (3) For low-resource languages, Machine Translation provides additional supervision and can be used as data augmentation. With the augmentation, our model improves mean recall by a large margin and obtains the highest scores on MSCOCO and Multi30k. In addition, the augmentation also improves mean recall on English.

\subsection{Comparison on Visual Features}
As shown in \cite{burnsLanguage2019}, EmbN is the state-of-the-art image-sentence model when using image-level ResNet features and good language features. However, prior work comparing cross-lingual image-sentence retrieval models only reported performance using VGG features.  These include VSE~\cite{kiros2014unifying}, Order Embeddings~\cite{vendrov}, PARALLEL-SYM/ASYM~\cite{gellaEMNLP2017}, and our implementation of EmbN~\cite{wang2019learning}. Table~\ref{tab:vgg} also shows the effect of going from VGG~\cite{simonyan2014very} to ResNet features.  As a reminder, we made some minor modifications to EmbN (see discussion in our paper), and we used the provided split for Multi30K, which is different than was used to benchmark EmbN on Flickr30K in~\cite{wang2019learning}.  Despite this, our reported results (mR 59.2) are quite comparable to the results in~\cite{wang2019learning} (mR 60.0).

\subsection{Non-English Languages on Multi30K}
From Table~\ref{tab:german} to Table~\ref{tab:czech}, the tables represent the performances of German, French, and Czech on Multi30K. For these three languages, our model (MULE EmbN + SA + LC + LP) outperforms the baselines, EmbN and PARALLEL-EmbN, by a large margin (German: Table~\ref{tab:german}(a) vs Table~\ref{tab:german}(d); French: Table~\ref{tab:french}(a) vs Table~\ref{tab:french}(d); Czech: Table~\ref{tab:czech}(a) vs Table~\ref{tab:czech}(d)). At the same time, our method is more scalable than others by using a universal embedding and a shared LSTM. The results show that our model transfer knowledge between languages and the performances of the low-resource languages are improved due to the alignment in languages. Especially for low-resource languages (\ie~French and Czech), the improvement is more significant than that of German. By comparing the number of (b, c) and (d) in Tables~\ref{tab:german},~\ref{tab:french},~\ref{tab:czech}, MUSE and multilingual BERT perform worse than our MULE. In addition, we observe that the low-resource languages can take benefit from Machine Translation. Comparing Table~\ref{tab:french}(a) and Table~\ref{tab:french}(e), baseline methods have similar performances to the model which is trained only on English and takes English translation sentences. The advantages of Machine Translation are more evident in Czech by comparing numbers in Table~\ref{tab:czech}(a) and Table~\ref{tab:czech}(f). Overall, we improve mean recall by 4.4\% for French and 7.1\% for Czech compared to the baseline PARALLEL-EmbN. After we augment the dataset with Machine Translation, we improve mean recall by 2.5\% for German, 10.3\% for French, and 14.5\% for Czech compared to the baseline PARALLEL-EmbN.  

\begin{table*}[h]
\centering
\setlength{\tabcolsep}{2.5pt}
\begin{tabular}{|rl|c|c|c|c|c|c|c|c|}
\hline
\cline{4-10}
& & Single & \multicolumn{3}{|c|}{Image-to-Sentence} & \multicolumn{3}{|c|}{Sentence-to-Image} & \\
\cline{4-10}
& Method & Model & R@1 & R@5 & R@10 & R@1 & R@5 & R@10 & mR \\
\hline
\hline
{\bf (a)} & \textbf{FastText (Baselines)} & & & & & & & & \\
& EmbN & N & 41.1 & 73.4 & 82.3 & 28.8 & 56.0 & 66.1 & 57.9 \\
& PARALLEL-EmbN & Y & 45.5 & 74.8 & 83.5 & 32.4 & 61.3 & 72.4 & 61.6\\
\hline
{\bf (b)} & \textbf{ML BERT} & & & & & & & & \\
& EmbN  & Y & 23.9 & 50.6 & 63.8 & 13.0 & 33.0 & 43.2 & 37.9\\
& PARALLEL-EmbN & Y & 35.1 & 62.8 & 74.6 & 23.7 & 49.7 & 61.0 & 51.1\\
\hline
{\bf (c)} & \textbf{MUSE} & & & & & & & & \\
& EmbN & Y & 42.4 & 71.7 & 82.2 & 29.2 & 56.6 & 67.3 & 58.2\\
& PARALLEL-EmbN & Y & 43.5 & 72.6 & 83.2 & 30.1 & 57.3 & 67.4 & 59.0\\
& EmbN + NC + LC + LP & Y & 44.8 & 73.6 & 83.3 & 30.2 & 58.3 & 67.8 & 59.7\\
\hline
{\bf (d)} &  \textbf{MULE (Ours)} & & & & & & & & \\
& EmbN & Y & 42.4 & 71.9 & 82.4 & 30.5 & 58.8 & 69.2 & 59.2\\
& EmbN + LC & Y & 45.5 & 71.8 & 83.7 & 32.9 & 60.7 & 71.1 & 61.0\\
& EmbN + LP & Y & 45.4 & 72.6 & 82.3 & 32.6 & 60.2 & 71.4 & 60.8\\
& EmbN + NC & Y & 42.4 & 73.2 & 82.2 & 31.3 & 59.6 & 71.0 & 59.9\\
& EmbN + NC + LP & Y & 44.3 & 72.8 & 83.2 & 32.1 & 62.1 & 72.0 & 61.1\\
& EmbN + LC + LP & Y & 44.0 & 74.3 & 85.0 & 32.5 & 60.9 & 70.9 & 61.3\\
& EmbN + NC + LC & Y & 45.2 & 73.5 & 82.8 & 31.3 & 60.8 & 71.8 & 60.9\\
& EmbN + NC + LC + LP (Full) & Y & 45.9 & 74.8 & 83.7 & 31.7 & 60.8 & 71.6 & 61.4\\
\hline
{\bf (e)} & \textbf{Translation to English} & & & & & & & & \\
& EmbN - English Model & Y & 34.1 & 60.4 & 71.1 & 19.6 & 47.4 & 58.5 & 48.5\\
\hline
{\bf (f)} & \textbf{Translation Data Augmentation} & & & & & & & & \\
& EmbN & N  & 46.6 & 73.9 & 82.2 & 31.3 & 59.1 & 69.0 & 60.3\\
& PARALLEL-EmbN & Y  & 46.1 & 76.3 & 83.2 & 34.4 & 62.5 & 73.0 & 62.6\\
& MULE EmbN - Full - Trans Only & Y & 46.9 & 74.6 & 82.3 & 34.0 & 62.2 & 72.7 & 62.1\\
& MULE EmbN - Full & Y & \textbf{49.7} & \textbf{77.7} &\textbf{85.7} & \textbf{34.6} & \textbf{63.4} & \textbf{73.5} & \textbf{64.1}\\
\hline
\end{tabular}
\caption{Image-sentence retrieval results on German. Our model outperforms the baselines by a large margin.}
\label{tab:german}
\end{table*}

\begin{table*}[h]
\centering
\setlength{\tabcolsep}{2.5pt}
\begin{tabular}{|rl|c|c|c|c|c|c|c|c|}
\hline
\cline{4-10}
& & Single  & \multicolumn{3}{|c|}{Image-to-Sentence} & \multicolumn{3}{|c|}{Sentence-to-Image} & \\
\cline{4-10}
& Method & Model & R@1 & R@5 & R@10 & R@1 & R@5 & R@10 & mR \\
\hline
\hline
{\bf (a)} & \textbf{FastText (Baselines)} & & & & & & & & \\
& EmbN & N & 23.6 & 46.2 & 57.4 & 24.7 & 49.4 & 59.3 & 43.4\\
& PARALLEL-EmbN & Y & 28.7 & 57.9 & 67.1 & 30.5 & 57.7 & 69.8 & 52.0\\
\hline
{\bf (b)} & \textbf{ML BERT} & & & & & & && \\
& EmbN & Y & 17.3 & 41.8 & 51.7 & 16.9 & 39.6 & 50.9 & 36.4\\
& PARALLEL-EmbN & Y & 20.9 & 46.8 & 57.3 & 21.8 & 47.0 & 58.1 & 42.0\\
\hline
{\bf (c)} & \textbf{MUSE} & & & & & & && \\
& EmbN & Y & 30.7 & 58.8 & 69.8 & 32.0 & 61.1 & 71.8 & 54.0\\
& PARALLEL-EmbN & Y & 29.3 & 57.0 & 67.6 & 28.3 & 58.3 & 69.3 & 51.6\\
& EmbN + NC + LC + LP & Y & 30.3 & 60.5 & 69.5 & 31.1 & 58.4 & 71.7 & 53.6\\
\hline
{\bf (d)} & \textbf{MULE (Ours)} & & & & & & && \\
& EmbN & Y & 30.0 & 57.1 & 68.0 & 30.1 & 59.6 & 70.2 & 52.5\\
& EmbN + LC & Y & 30.0 & 58.2 & 69.2 & 31.6 & 59.8 & 70.3 & 53.2\\
& EmbN + LP & Y & 30.4 & 58.1 & 69.8 & 29.9 & 60.6 & 71.2 & 53.3\\
& EmbN + NC & Y & 30.7 & 58.8 & 71.5 & 31.3 & 59.6 & 71.0 & 54.0\\
& EmbN + NC + LP & Y & 31.6 & 59.8 & 70.9 & 30.3 & 59.8 & 71.7 & 54.0\\
& EmbN + LC + LP & Y & 30.4 & 58.4 & 68.3 & 31.0 & 59.1 & 70.4 & 52.9\\
& EmbN + NC + LC & Y & 30.8 & 61.7 & 73.2 & 31.7 & 62.5 & 72.8 & 55.5\\
& EmbN + NC + LC + LP (Full) & Y & 32.6 & 62.6 & 73.6 & 32.7 & 62.4 & 74.5 & 56.4\\
\hline
{\bf (e)} & \textbf{Translation to English} & & & & & && & \\
& EmbN - English Model & Y & 22.5 & 52.5 & 63.0 & 25.1 & 53.1 & 63.9 & 46.7\\
\hline
{\bf (f)} & \textbf{Translation Data Augmentation} & & & & & & && \\
& EmbN & N & 31.0 & 60.4 & 71.0 & 35.2 & 60.3 & 70.8 & 54.8\\
& PARALLEL-EmbN & Y & 37.6 & 66.0 & 77.4 & 37.8 & 66.4 & 78.2 & 60.6\\
& MULE EmbN - Full - Trans Only & Y & 35.8 &  \textbf{69.4} & 78.3 & \textbf{38.7} & 68.3 & 78.6 & 61.5\\
& MULE EmbN - Full & Y & \textbf{38.0} & 68.4 & \textbf{80.0} & 38.2 & \textbf{68.9} & \textbf{80.3} & \textbf{62.3}\\
\hline
\end{tabular}
\caption{Image-sentence retrieval results on French. Our model outperforms the baselines by a large margin. The model taking English translations from French (e) achieves similar performances to the baselines (a).}
\label{tab:french}
\end{table*}

\begin{table*}[h]
\centering
\setlength{\tabcolsep}{2.5pt}
\begin{tabular}{|rl|c|c|c|c|c|c|c|c|}
\hline
\cline{4-10}
& & Single  & \multicolumn{3}{|c|}{Image-to-Sentence} & \multicolumn{3}{|c|}{Sentence-to-Image} & \\
\cline{4-10}
& Method & Model & R@1 & R@5 & R@10 & R@1 & R@5 & R@10 & mR \\
\hline
\hline
{\bf (a)} & \textbf{FastText (Baselines)} & & & & & & & & \\
& EmbN & N & 16.6 & 34.9 & 45.0 & 17.7 & 37.7 & 48.6 & 33.4\\
& PARALLEL-EmbN & Y & 22.1 & 46.9 & 58.5 & 22.4 & 49.8 & 59.5 & 43.2\\
\hline
{\bf (b)} & \textbf{ML BERT} & & & & & & & &\\
& EmbN & Y & 8.0 & 22.0 & 33.4 & 6.1 & 18.5 & 26.9 & 19.2\\
& PARALLEL-EmbN & Y & 13.3 & 32.8 & 42.6 & 12.7 & 33.0 & 44.7 & 29.8\\
\hline
{\bf (c)} & \textbf{MUSE} & & & & & & & &\\
& EmbN & Y & 23.3 & 45.6 & 55.3 & 21.9  &47.6 & 57.4 & 41.8\\
& PARALLEL-EmbN & Y & 21.0 & 44.7 & 55.2 & 23.5 & 44.4 & 55.6 & 40.7\\
& EmbN + NC + LC + LP & Y & 21.3 & 44.1 & 54.2 & 22.9 & 46.3 & 57.0 & 41.0\\
\hline
{\bf (d)} & \textbf{MULE (Ours)} & & & & & & & & \\
& EmbN & Y & 23.0 & 47.3 & 59.4 & 23.8 & 48.9 & 61.1 & 43.9\\
& EmbN + LC & Y & 23.9 & 47.8 & 59.6 & 24.5 & 50.3 & 61.1 & 44.5\\
& EmbN + LP & Y & 23.5 & 48.1 & 59.1 & 24.7 & 49.2 & 60.3 & 44.1\\
& EmbN + NC & Y & 23.7 & 50.2 & 63.3 & 25.6 & 51.6 & 64.1 & 46.4\\
& EmbN + NC + LP & Y & 25.8 & 54.3 & 66.3 & 28.6 & 54.1 & 65.4 & 49.1\\
& EmbN + LC + LP & Y & 22.7 & 48.9 & 59.3 & 25.4 & 51.0 & 60.6 & 44.6\\
& EmbN + NC + LC & Y & 25.6 & 54.2 & 66.7 & 27.4 & 56.4 & 66.5 & 49.5\\
& EmbN + NC + LC + LP (Full) & Y & 26.6 & 56.4 & 67.2 & 27.8 & 56.1 & 67.5 & 50.3\\
\hline
{\bf (e)} & \textbf{Translation to English} & & & & & & & & \\
& EmbN - English Model & Y & 23.0 & 50.9 & 64.7 & 25.1 & 53.4 & 64.2 & 46.9\\
\hline
{\bf (f)} & \textbf{Translation Data Augmentation} & & & & & & & & \\
& EmbN & N & 26.2 & 51.3 & 62.5 & 26.8 & 50.3 & 60.8 & 46.3\\
& PARALLEL-EmbN & Y & 31.4 & 58.2 & 70.1 & 33.1 & 60.4 & 71.6 & 54.1\\
& MULE EmbN - Full - Trans Only & Y & 33.1 & 60.0 & 71.3 & 33.5 & 62.4 & 72.9 & 55.5\\
& MULE EmbN - Full & Y & \textbf{34.3} & \textbf{63.2} & \textbf{74.2} & \textbf{35.3} & \textbf{63.6} & \textbf{75.5} & \textbf{57.7}\\
\hline
\end{tabular}
\caption{Image-sentence retrieval results on Czech. Our model outperforms the baselines by a large margin. The model taking English translations from Czech (e) outperforms to the baselines (a).}
\label{tab:czech}
\end{table*}

\subsection{Non-English Languages on MSCOCO}
We see similar behavior on MSCOCO that our method performs better on low-resource languages than the baselines. The big difference from Multi30K is that Machine Translation significantly improves performances on Chinese as shown in Table~\ref{tab:chinese} (d). This could be due to the fact that the number of Chinese annotations is much less than that of other languages. Based on the observation, we augment the dataset with Machine Translation. We show that our model with the augmentation obtain the highest scores as shown in Table~\ref{tab:chinese} (e) and Table~\ref{tab:japanese} (e). Our approach with the data augmentation improves mean recall by 21.9\% for Chinese and 7.4\% for Japanese compared to the baseline PARALLEL-EmbN.

\begin{table*}[h]
\centering
\setlength{\tabcolsep}{2.5pt}
\begin{tabular}{|rl|c|c|c|c|c|c|c|c|}
\hline
\cline{4-10}
& & Single & \multicolumn{3}{|c|}{Image-to-Sentence} & \multicolumn{3}{|c|}{Sentence-to-Image} & \\
\cline{4-10}
& Method & Model  & R@1 & R@5 & R@10 & R@1 & R@5 & R@10 & mR \\
\hline
\hline
{\bf (a)} & \textbf{FastText (Baselines)} & & & & & & & &\\
& EmbN & N & 29.1 & 61.4 & 74.1 & 30.0 & 64.8 & 74.8 & 55.7\\
& PARALLEL-EmbN  & Y & 28.6 & 58.4 & 71.7 & 28.5 & 58.4 & 71.8 & 52.9\\
\hline
{\bf (b)} & \textbf{ML BERT} & & & & & & && \\
& EmbN & Y & 22.1 & 53.5 & 66.5 & 20.0 & 45.7 & 60.4 & 44.7\\
& PARALLEL-EmbN & Y & 29.3 & 62.1 & 77.0 & 30.6 & 65.1 & 80.9 & 57.5\\
\hline
{\bf (c)} & \textbf{MULE (Ours)} & & & & & & && \\
& EmbN & Y & 29.9 & 60.4 & 73.0 & 30.0 & 61.4 & 70.4 & 54.2\\
& EmbN + LC & Y & 28.8 & 60.0 & 70.6 & 29.9 & 63.3 & 75.3 & 54.7\\
& EmbN + LP & Y & 33.8 & 63.9 & 75.7 & 33.9 & 64.4 & 75.4 & 57.9\\
& EmbN + NC & Y & 28.7 & 62.2 & 74.4 & 34.2 & 65.5 & 74.5 & 56.6\\
& EmbN + NC + LP & Y & 32.7 & 63.6 & 76.1& 33.6 & 63.4 & 73.8 & 57.2\\
& EmbN + LC + LP & Y & 32.2& 65.2 & 77.3 & 33.0 & 64.1 & 77.3 & 58.2\\
& EmbN + NC + LC & Y & 32.9 & 63.9 & 76.7 & 31.7 & 65.3 & 76.7 & 57.9\\
& EmbN + NC + LC + LP (Full) & Y & 34.4 & 60.2 & 75.8 & 34.9 & 66.0 & 77.2 & 58.8\\
\hline
{\bf (d)} & \textbf{Translation to English} & & & & & & && \\
& EmbN - English Model & Y & 45.9 & 79.8 & 89.2 & 47.8 & 81.1 & 89.4 & 72.2\\
\hline
{\bf (e)} & \textbf{Translation Data Augmentation} & & & && & & & \\
& EmbN & N & 49.6 & 81.6 & 90.0 & 47.8 & 82.1 & 90.0 & 73.5\\
& PARALLEL-EmbN & Y & 47.9 & 81.4 & 91.1 & 47.5 & 81.6 & 91.2 & 73.5\\
& MULE EmbN - Full - Trans Only & Y & 49.1 & 80.8 & 90.8 & 48.0 & 80.9 & 90.2 & 73.3\\
& MULE EmbN - Full & Y & \textbf{51.1} & \textbf{82.6} & \textbf{91.6} & \textbf{49.1} & \textbf{82.4} & \textbf{91.9} & \textbf{74.8}\\
\hline
\end{tabular}
\caption{Image-sentence retrieval results on Chinese. By comparing numbers in (a, b) and (c), our model achieves the best performances compared to the other baselines. For Chinese, Machine Translation significantly boosts the performances as shown in (d). Based on the observation, we augment the training set with Machine Translation and show the data augmentation effectively works for Chinese. Our model obtains the best performances with the Machine Translation augmentation.}
\label{tab:chinese}
\end{table*}

\begin{table*}[h]
\centering
\setlength{\tabcolsep}{2.5pt}
\begin{tabular}{|rl|c|c|c|c|c|c|c|c|}
\hline
\cline{4-10}
& & Single & \multicolumn{3}{|c|}{Image-to-Sentence} & \multicolumn{3}{|c|}{Sentence-to-Image} & \\
\cline{3-9}
& Method & Model & R@1 & R@5 & R@10 & R@1 & R@5 & R@10 & mR \\
\hline
\hline
{\bf (a)} & \textbf{FastText (Baselines)} & && & & & & & \\
& EmbN & N & 47.6 & 81.4 & 89.9 & 39.1 & 73.2 & 85.4 & 69.4\\
& PARALLEL-EmbN  & Y & 49.6 & 79.7 & 90.1 & 39.0 & 71.9 & 83.2 & 68.9\\
\hline
{\bf (b)} & \textbf{ML BERT} & & & & & && & \\
& EmbN & Y& 26.5 & 60.3 & 75.8 & 18.1 & 44.2 & 58.0 & 47.2\\
& PARALLEL-EmbN & Y & 40.5 & 73.5 & 85.8 & 30.4 & 64.1 & 78.4 & 62.1\\
\hline
{\bf (c)} & \textbf{MULE (Ours)} & & & & && & & \\
& EmbN & Y & 50.4 & 80.2 & 89.4 & 38.8 & 71.8 & 83.7 & 69.0\\
& EmbN + LC & Y & 49.9 & 80.3 & 89.9 & 39.5 & 73.5 & 85.2 & 69.6\\
& EmbN + LP & Y & 49.9 & 81.0 & 90.1 & 38.6 & 72.8 & 83.5 & 69.3\\
& EmbN + NC & Y & 47.3 & 80.4 & 90.7 & 39.3 & 73.9 & 85.5 & 69.5\\
& EmbN + NC + LP & Y & 49.4 & 80.5 & 91.4 & 40.3 & 74.2 & 85.4 & 70.2\\
& EmbN + LC + LP & Y & 49.0 & 82.6 & 91.7 & 40.0 & 72.8 & 84.8 & 70.0\\
& EmbN + NC + LC & Y & 49.8 & 81.5 & 91.6 & 40.1 & 73.6 & 85.4 & 70.3\\
& EmbN + NC + LC + LP (Full) & Y & 49.9 & 81.4 & 92.0 & 40.4 & 73.8 & 85.5 & 70.5\\
\hline
{\bf (d)} & \textbf{Translation to English} & && & & & & & \\
& EmbN - English Model & Y & 44.8 & 74.3 & 85.4 & 36.9 & 71.0 & 84.7 & 66.1\\
\hline
{\bf (e)} & \textbf{Translation Data Augmentation} & && & & & & & \\
& EmbN & N & 56.0 & 83.7 & 90.7 & 45.5 & 77.2 & 87.3 & 73.2\\
& PARALLEL-EmbN & Y & \textbf{60.1} & 86.0 & 92.8 & 47.7 & 79.6 & 89.7 & 76.0\\
& MULE EmbN - Full - Trans Only & Y & 58.5& 85.6 & 92.6 & 46.1 & 79.2 & 89.8 & 75.3\\
& MULE EmbN - Full & Y & 59.6 & \textbf{86.5} & \textbf{92.8} & \textbf{47.8} & \textbf{80.8} & \textbf{90.1} & \textbf{76.3}\\
\hline
\end{tabular}
\caption{Image-sentence retrieval results on Japanese. By comparing numbers in (a, b) and (c), our model achieves the best performances compared to the other baselines. From (e), we observe that the data augmentation with Machine Translation improves performances by a large margin. Our method achieves the highest score with the data augmentation with Machine Translation.}
\label{tab:japanese}
\end{table*}

\subsection{English on Multi30K and MSCOCO}
Table~\ref{tab:english} shows the overall results on English. The monolingual model (EmbN) performs better on English than multilingual models on Multi30K, while our method outperforms the monolingual model on MSCOCO. From Table~\ref{tab:english}(e), we observe that Machine translation also improves the performances on English.

\begin{table*}[h]

\centering
\setlength{\tabcolsep}{1.5pt}
\begin{tabular}{|rl|c|c|c|c|c|c|c|c|c|c|c|c|c|c|c|}
\hline
& & & \multicolumn{7}{|c|}{Multi30K} & \multicolumn{7}{|c|}{MSCOCO}\\
\cline{4-17}
& & Single & \multicolumn{3}{|c|}{Image-to-Sentence} & \multicolumn{3}{|c|}{Sentence-to-Image} & & \multicolumn{3}{|c|}{Image-to-Sentence} & \multicolumn{3}{|c|}{Sentence-to-Image} &\\
\cline{4-17}
& Method & Model & R@1 & R@5 & R@10 & R@1 & R@5 & R@10 & mR & R@1 & R@5 & R@10 & R@1 & R@5 & R@10 & mR\\
\hline
\hline
{\bf (a)} & \textbf{FastText (Baselines)} & & & & & & & & & & & & & & & \\
& EmbN & N & \textbf{58.3} & 82.9 & 90.4 & 41.7 & 72.0 & 81.2 & 71.1 & 58.6 & 86.5 & 94.1 & 45.5 & 79.6 & 89.5 & 75.6\\
& PARALLEL-EmbN & Y & 55.9 & 82.5 &  88.4 & 40.3 & 70.1 & 80.2 & 69.6 & 52.3 & 82.9 & 90.3 & 39.2 & 72.0 & 83.5 & 70.0\\
\hline
{\bf (b)} & \textbf{ML BERT} & & & & & & & & & & & & & & &\\
& EmbN & Y  & 28.6 & 58.0 & 69.5 & 18.7 & 43.6 & 54.4 & 45.5 & 36.7 & 70.4 & 82.7 & 28.4 & 61.9 & 76.4 & 59.4\\
& PARALLEL-EmbN & Y & 42.4 & 74.3 & 83.1 & 30.3 & 60.4 & 71.7 & 60.4 & 30.9 & 65.8 & 81.1 & 28.0 & 62.7 & 77.4 & 57.6\\
\hline
{\bf (c)} & \textbf{MUSE} & & & & & & & & & & & & & & &\\
& EmbN & Y & 53.6 & 82.4 & 88.6 & 38.1 & 69.7 & 79.1 & 68.6 & -- & -- & -- & -- & -- & -- & --\\
& PARALLEL-EmbN & Y & 57.5 & 81.6 & 89.2 & 39.6 & 69.7 & 79.6 & 69.5 & -- & -- & -- & -- & -- & -- & --\\
& EmbN NC + LC + LP & Y & 55.1 & 81.6 & 88.2 & 39.7 & 69.9 & 79.7 & 69.0 & -- & -- & -- & -- & -- & -- & --\\
\hline
{\bf (d)} & \textbf{MULE (Ours)} & & & & & & & & & & & & & & &\\
& EmbN & Y & 52.1 & 79.6 & 87.5 & 37.8 & 68.9 & 79.0 & 67.5 & 50.7 & 83.1 & 90.6 & 38.3 & 71.1 & 82.4 & 69.4\\
& EmbN + LC & Y & 52.7 & 79.6 & 87.4 & 38.5 & 68.8 & 78.8 & 67.6 & 51.1 & 83.2 & 92.5 & 39.1 & 72.0 & 82.3 & 70.0\\
& EmbN + LP & Y & 51.8 & 79.8 & 87.9 & 38.7 & 69.2 & 78.6 & 67.7 & 54.3 & 83.1 & 89.6 & 37.9 & 72.9 & 84.1 & 70.3\\
& EmbN + NC & Y & 49.4 & 79.1 & 87.0 & 38.6 & 69.4 & 79.6 & 67.2 & 51.1 & 83.5 & 91.0 & 38.7 & 71.5& 82.8 & 69.8\\
& EmbN + NC + LP & Y & 53.0 & 79.0 & 87.3 & 38.2 & 69.5 & 79.4 & 67.7 & 52.3 & 81.8 & 91.6 & 38.3 & 73.0 & 84.2 & 70.2\\
& EmbN + LC + LP & Y & 52.2 & 80.5 & 88.4 & 38.7 & 69.1 & 79.4 & 68.1 & 54.0 & 82.6 & 90.0 & 36.5 & 72.7 & 84.0 & 70.0\\
& EmbN + NC + LC & Y & 49.2 & 80.5 & 87.3 & 37.8 & 68.7 & 79.3 & 67.1 & 51.3 & 83.0 & 92.9 & 40.1 & 74.2 & 86.2 & 71.3\\
& EmbN + NC + LC + LP (Full) & Y & 52.3 & 79.8 & 87.8 & 38.4 & 69.5 & 79.9 & 68.0 & 53.5 & 83.1 & 92.9 & 42.1 & 74.4 & 85.9 & 72.0\\
\hline
{\bf (e)} & \textbf{Translation Data Augmentation} & & & & & & & & & & & & & & &\\
& EmbN & N & \textbf{57.9} &  \textbf{84.5} & \textbf{90.9} & \textbf{44.3} & \textbf{72.7} & \textbf{84.7} & \textbf{72.0} & 61.8 & 87.6 & 94.1 & 47.5 & 79.8 & 89.8 & 76.8\\
& PARALLEL-EmbN & Y & 52.4 & 80.1 & 87.7 & 41.6 & 71.5 & 80.7 & 69.0 & 63.1 & 89.1 & 94.1 & 49.2 & 82.5 & 91.5 & 78.3\\
& MULE EmbN - Full - Trans Only & Y & 54.2 & 80.7 & 87.8 & 41.3 & 71.2 & 80.4 & 69.3 & 60.8 & 88.1 & 94.1 & 47.9 & 81.6 & 91.0 & 77.3\\
& MULE EmbN - Full & Y & 54.2 & 82.0 & 89.9 & 41.9 & 72.5 & 81.1 & 70.3 & \textbf{63.9} & \textbf{90.2} & \textbf{95.8} & \textbf{50.9} & \textbf{83.5} & \textbf{92.4} & \textbf{79.5}\\
\hline
\end{tabular}
\caption{Image-sentence retrieval results on English. For the label-rich language, English, the monolingual model (EmbN) outperforms multilingual models on Multi30K and MSCOCO.}
\label{tab:english}
\end{table*}



\end{document}